\documentclass{article}

\usepackage{amssymb, amsmath, amsopn}
\usepackage{times,mathptmx}

 \usepackage{mathptmx}
 \usepackage{helvet}
 \usepackage{courier}
 \usepackage{makeidx}
 \usepackage{multicol}
 \usepackage{footmisc}
 \usepackage[justification=centering]{caption}
 \usepackage{amssymb, amsmath}
 \usepackage{graphicx}
 \usepackage{caption}
 \usepackage{subfig}
 \usepackage{xspace}
 \usepackage{color}
 \usepackage{array}
 \newcolumntype{P}[1]{>{\centering\arraybackslash}p{#1}}
 \usepackage{verbatim}
 \usepackage{imakeidx, epsfig,url}
\usepackage{algorithm}
\usepackage[algo2e,ruled,vlined]{algorithm2e}
\usepackage{xspace}
\usepackage[numbers,sort&compress,longnamesfirst,sectionbib]{natbib}
\usepackage{balance}

\usepackage{graphicx}

\begin{document}

\title{A Feature-Based Analysis on the Impact of Set of Constraints for $\epsilon$-Constrained Differential Evolution}
\author{
Shayan Poursoltan\\
Optimisation and Logistics\\
School of Computer Science\\
The University of Adelaide\\
Adelaide, Australia
\and 
 Frank Neumann\\
 Optimisation and Logistics\\
 School of Computer Science\\
 The University of Adelaide\\
 Adelaide, Australia
}

\maketitle
\begin{abstract}
	Different types of evolutionary algorithms have been developed for constrained continuous optimization. 
	We carry out a feature-based analysis of evolved constrained continuous optimization instances to understand the characteristics of constraints that make problems hard for evolutionary algorithm. In our study, we examine how various sets of constraints can influence the behaviour of $\epsilon$-Constrained Differential Evolution. Investigating the evolved instances, we obtain knowledge of what type of constraints and their features make a problem difficult for the examined algorithm.
\end{abstract}

\section{Introduction}
Constrained optimisation problems (COPs), specially non-linear ones, are very important and widespread in real world applications \cite{floudas1990collection}. This has motivated introducing various algorithms to solve COPs. The focus of these algorithms is to handle the involved constraints. In order to deal with the constraints, various mechanisms have been adopted by evolutionary algorithms. These techniques include penalty function, decoder-based methods and special operators that separate the treatment of constraints and objective functions. For an overview of different types of methods we refer the reader to Mezura-Montes and Coello Coello \cite{mezura2011constraint}.

With the increasing number of evolutionary algorithms, it is hard to predict which algorithm performs better for a newly given COP. 	
Various benchmark sets such as CEC'10 \cite{mallipeddi2010problem} and BBOB'10 \cite{hansen2010real} have been proposed to evaluate the algorithm performances on continuous optimization problems. The aim of these benchmarks is to find out which algorithm is good on which classes of problems.
For constrained continuous optimization problems, there has been an increasing interest to understanding problem features from a theoretical perspective \cite{poursoltan2015ruggedness,vassilev2000information}. 
 The feature-based analysis of of hardness for certain classes of algorithms is a relatively new research area. Such studies classify problems as hard or easy for a given algorithm based on the features of given instances. Initial studies in the context of continuous optimization have  recently been carried out in~\cite{DBLP:conf/ppsn/MersmannPT10,mersmann2011exploratory}.
	Having enough knowledge on problem properties that make it hard or easy, we may choose the most suited algorithm to solve it. To do this, two steps approach has been proposed by Mersmann et al.~\cite{mersmann2011exploratory}. First, one has to extract the important features from a group of investigated problems. Second, in order to build a prediction model, it is necessary to analyse the performance of various algorithms on these features. 
 Feature-based analysis has also been used to gain new insights in algorithm performance for discrete optimization problems \cite{smith2010understanding,nallaperuma2013feature}.
	
In this paper, we carry out a feature-based analysis for constrained continuous optimisation and generate a variety of problem instances from easy to hard ones by evolving constraints. This ensures that the knowledge obtained by analysing problem features covers a wide range of problem instances that are of particular interest. Although what makes a problem hard to solve is not a standalone feature, it is assumed that constraints are certainly important in constrained problems. Evolving constraints is a new technique to generate hard and easy instances. 
So far, the influence of one linear constraint has been studied \cite{poursoltan2014feature}. However, real world problems have more than one linear constraint (such as linear, quadratic and their combination). Hence, our study is to generate COP instances to investigate which features of the linear and quadratic constraints make the constrained problem hard to solve. To provide this knowledge, we need to use a common suitable evolutionary algorithm that handles the constraints. In this research, the $\epsilon$-constrained differential evolution with an archive and gradient-based mutation ($\epsilon$DEag) \cite{takahama2010constrained} is used. The $\epsilon$DEag (winner of CEC 10 special session for constrained problems) is applied to generate hard and easy instances to analyse the impact of set of constraints on it.
	
	Our results provide evidence on the capability of constraints (linear, quadratic or their set of combination) features to classify problem instances to easy and hard ones. Feature analysis by solving the generated instances with $\epsilon$DEag enables us to obtain the knowledge of influence of constraints on problem hardness which could later could be used to design a successful prediction model for algorithm selection. 
	
	The rest of the paper is organised as follows. In Section 2, we introduce the constrained optimisation problems. Then, we discuss  $\epsilon$DEag algorithm that we use to solve the generated problem instances. Section 3 includes our approach to evolve and generate problem instances. Furthermore, the constraint features are discussed. In Section 4, we carry out the analysis of the linear and quadratic constraint features. Finally, section 5 concludes with some remarks.

\section{Preliminaries}
\subsection{Constrained continuous optimisation problems}
Constrained continuous optimisation problems are optimisation problems where a function $f(x)$ on real-valued variables should be optimised with respect to a given set of constraints. Constraints are usually given by a set of inequalities and/or equalities. Without loss of generality, we present our approach for minimization problems.

Formally, we consider single-objective functions $f \colon S \rightarrow \mathbb{R}$, with $S \subseteq \mathbb{R}^n$.
The constraints impose a feasible subset $F \subseteq S$ of the search space $S$ and the goal is to find an element $x \in S \cap F$ that minimizes $f$.

We consider problems of the following form:

		\begin{equation}
		\begin{split}
		\text{minimize } &\quad    f(x), \quad	x = (x_1,\ldots,x_n) \in \mathbb{R}^n  \\
		\text{subject to}&\quad	g_i(x)  \leq 0 \quad  \forall   i \in \{1,\ldots,q\}\\
		&	\quad	h_j(x) = 0 \quad \forall j \in \{q + 1,\ldots, p\} 
		\label{eq:f}
		\end{split}
		\end{equation}

 where $x = (x_1,x_2,\dots,x_n)$ is an $n$ dimensional vector and $x \in S  \cap F$. Also $g_{i}(x)$ and $h_{j}(x)$ are inequality and equality constraints respectively. Both inequality and equality constraints could be linear or nonlinear. To handle equality constraints, they are usually transformed into inequality constraints as 	$|h_j(x)|  \leq \epsilon$, where $\epsilon=10e^{-4}$ (used in \cite{mallipeddi2010problem}). Also, the feasible region $F \subseteq S$ of the search space $S$ is defined by
\begin{equation}
l_i  \leq x_i  \leq u_i,\quad  \quad     1 \leq i  \leq n
\label{equality}
\end{equation}

 where both $l_i$  and $u_i$ denote lower and upper bounds for the $i$th variable and $1\leq i \leq n$ respectively.

\subsection{$\epsilon$DEag algorithm }
One of the most prominent evolutionary algorithms for COPs is $\epsilon$-constrained differential evolution with an archive and gradient-based mutation ($\epsilon$DEag). The algorithm is the winner of latest CEC competition for constrained constrained continuous problems \cite{mallipeddi2010problem}. The $\epsilon$DEag uses $\epsilon$-constrained method to transform algorithms for unconstrained problems to constrained ones \cite{takahama2008constrained}. It adopts $\epsilon$-level comparison instead of ordinary ones to order the possible solutions. In other words, the lexicographic order is performed in which constraint violation ($\phi(x)$) has more priority and proceeds the function value ($f(x)$). This means feasibility is more important. Let $f_{1}$,$f_{2}$ and $\phi_{1}$,$\phi_{2}$ are objective function values and constraint violation at $x_{1}$,$x_{2}$ respectively. Hence, for all $\epsilon \geq 0$, the $\epsilon$-level comparison of two candidates $(f_{1},\phi_{1})$ and $(f_{2},\phi_{2})$ is defined as the follows:

		\begin{equation*}
		\label{eq:StringS}
		(f_{1},\phi_{1}) <_{\epsilon}  (f_{2},\phi_{2}) \iff   \begin{cases}
		f_{1} < f_{2}, & \text{\quad if} \quad \phi_{1},\phi_{2} \leq \epsilon \\
		f_{1} < f_{2}, & \text{\quad if} \quad \phi_{1} = \phi_{2} \\
		\phi_{1} < \phi_{2}, & \quad  \text{otherwise} \\
		\end{cases} 
		\end{equation*}
		\begin{equation*}
		\label{eq:StringS}
		(f_{1},\phi_{1}) \leq_{\epsilon}  (f_{2},\phi_{2}) \iff   \begin{cases}
		f_{1} \leq f_{2}, & \text{\quad if} \quad \phi_{1},\phi_{2} \leq \epsilon \\
		f_{1} \leq f_{2}, & \text{\quad if} \quad \phi_{1} = \phi_{2} \\
		\phi_{1} < \phi_{2}, & \quad  \text{otherwise} \\
		\end{cases} 
		\end{equation*}

In order to improve the usability, efficiency and stability of the algorithm, an archive has been applied. Using it improves the diversity of individuals (see Algorithm \ref{alg:edeag}). The offspring generation is adopted in such a way that if the child is not better than its parent, the parent generates another one (see \cite{takahama2010constrained}). This leads to more stability to the algorithm. 
For a detailed presentation of the algorithm, we refer the reader to \cite{takahama2010constrained}.

\begin{algorithm}[t]
	
	\begin{itemize}
		\itemsep-0.1em
		\item Initializations:\\
		- \textit{M} randomly selected individuals from search space \textit{S} is archived in \textit{A}.\\
		- Set $\epsilon$ level Using control level function\\
		- Population: Top \textit{N} individuals are selected from archive. The Archive is ranked using $\epsilon$ level comparison
		\item Termination condition is set to Maximum function evaluation number.
		\item DE operation: Use DE/rand/1/exp to generate new child. Comparing is based on the $\epsilon$ level comparison
		\item Gradient based mutation: If child is infeasible, it is changed by the gradient-based mutation with probability P. Go to step 3 and parent is considered as parent. 
		\item Update and control the $\epsilon$-level
		\item Go to step 2
		
	\end{itemize}
	
	\caption{The $\epsilon$-constrained differential evolution with an archive and gradient-based mutation ($\epsilon$DEag)}
	
	\label{alg:edeag}
\end{algorithm}

\section{Evolving Constraints}
It is assumed that the role of constraints in problem difficulty is certainly important for constrained optimisation problem. Hence, it is necessary to analyse various effects that constraint can impose on a constrained optimisation problems. Evolving constraints is a novel methodology to generate hard and easy instances based on the performance of the problem solver (optimisation algorithm).

\subsection{Algorithm}
\begin{figure}[t]
		
		\centering
		\includegraphics[scale=0.8]{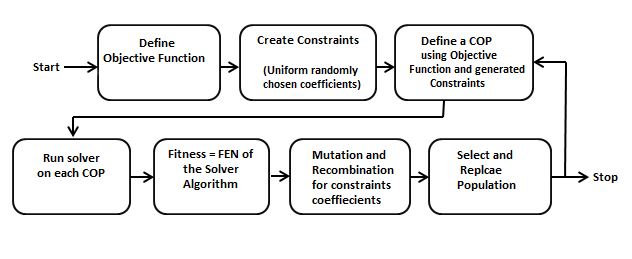}
		\caption{Evolving constraints process}
		\label{fig:flowchart}
	\end{figure}

In order to analyse the effects of constraints, the variety of them needs to be studied over a fixed objective function. First, constraint coefficients are randomly chosen to construct problem instances. Second, the generated constrained problem is solved by a solver algorithm ($\epsilon$DEag). Then, the required function evaluation number (FEN) to solve this instance is considered as the fitness value for evolving algorithm. This process is repeated until hard and easy instances of constraint problem are generated (see Figure \ref{fig:flowchart}).

To generate hard and easy instances, we use the approach outlined in \cite{poursoltan2014feature}. 

It uses fast and robust differential evolution (DE) proposed in \cite{storn1997differential} (see Algorithm \ref{alg:DE}, \ref{alg:newsample}) to evolve through the problem instances (by generating various constraint coefficients). It is necessary to note that the aim is to optimise (maximise/minimise) the FEN that is required by a solver to solve the generated problem. 
Also, to solve this generated problem instance and find the required FEN we use $\epsilon$DEg as a solver. 
The termination condition of this algorithm (evolver) is set to reaching FENmax number of function evaluations or finding a solution close enough to the feasible optimum solution as follows:

	\begin{equation}
	\label{eq:fenmax}
	|f(x_{optimum})-f(x_{best})| \leq e^{-12} 
	\end{equation}

			\begin{algorithm}[t]
				
				\begin{itemize}
					\itemsep-0.3em
					\item inputs: Problem and $Pop_{size}$, $Crossover_{rate}$ ,$weighting_{factor}$, outputs: $S_{best}$ 
					
					\item Population $\leftarrow$ InitializePop\\ EvaluatePopulation(population)\\
					$S_{best}$ $\leftarrow$ GetBestSolution(Population)
					\item \textbf{Repeat}
					\item\quad  NewPopulation $\leftarrow$ $\phi$
					\item \quad \textbf{For} i starts at 1, i$<Pop_{size}$-1, increment i
					\item \quad \quad $S_i$ $\leftarrow$ Newsample
					\item \quad \quad \textbf{If} Cost($S_i$)$\le$Cost($P_{i}$)
					\item \quad \quad \quad NewPopulation $\leftarrow$ $S_i$
					\item \quad \quad \textbf{else}
					\item \quad \quad \quad NewPopulation $\leftarrow$ $P_i$
					\item \quad \quad \textbf{Endif}
					\item \quad \textbf{ Endfor}
					\item \quad Population $\leftarrow$ NewPopulation
					\item \quad EvaluatePopulation(population)
					\item \quad $S_{best}$ $\leftarrow$ GetBestSolution(Population)
					\item \textbf{Until} (stop condition)
				\end{itemize}
				\caption{Differential evolution (DE) algorithm}
				\label{alg:DE}
			\end{algorithm}

			\begin{algorithm}[t]
			
				\begin{itemize}
					\itemsep-0.3em
					\item inputs: $P_{0}$, population, NP, F, CR, outputs: $S$ 
					\item \textbf{Repeat}
					\item \quad $P_{1}$ $\leftarrow$ RandomMember(population)
					\item \textbf{Untill} $P_{1}$ $\not=$ $P_{1}$
					\item \textbf{Repeat}
					\item \quad $P_{2}$ $\leftarrow$ RandomMember(population)
					\item \textbf{Untill} $P_{2}$ $\not=$ $P_{0}$ $\vee$  $P_{2}$ $\not=$ $P_{1}$
					\item \textbf{Repeat}
					\item \quad $P_{3}$ $\leftarrow$ RandomMember(population)
					\item \textbf{Untill} $P_{3}$ $\not=$ $P_{0}$ $\vee$  $P_{3}$ $\not=$ $P_{1}$  $P_{3}$ $\not=$ $P_{2}$ 
					\item cutpoint $\leftarrow$ RandomMember(population)
					\item $Sample$ $\leftarrow$ 0
					\item \textbf{For} i starts a 1 to NP
					\item \quad \textbf{If} i $\equiv$ cutpoint $\wedge$ Rand() $\leq$ CR
					\item \quad \quad $S_{i}$ $\leftarrow$ $P_{3_{i}}$ + F*($P_{1_{i}}$-$P_{2_{i}}$)
					\item \quad \textbf{Else}
					\item \quad \quad $S_{i}$ $\leftarrow$ $P_{0_{i}}$
					\item \quad \textbf{Endif}
					\item \quad \textbf{Endfor}
					\item Return $S$
					
				\end{itemize}
				\caption{ Newsample function in Algorithm  \ref{alg:DE}}
				\label{alg:newsample}
			\end{algorithm}

This process generates harder and easier problem instances until it reaches the certain number of generation for the DE algorithm (evolver). Once two distinct sets of easy and hard instances are ready, we start analysing various features of the constraints for these two categories. This could give us the knowledge to understand which features of constraints have more contribution to problem difficulty.

\subsection{Evolving a set of inequality constraints}
We focus on analysing the effects of constraints (linear, quadratic and their combination) on the problem and algorithm difficulty. We will extract features of constraints and analyse their effect on  constrained problem difficulty. The experimented constraints are linear and quadratic as the form of: \\

		\begin{equation}
		\text{linear constraint} \quad	g(x) = b + a_{1}x_{1} + \ldots + a_{n}x_n
		\label{lineareq}
		\end{equation}

		\begin{equation}
			\text{quadratic constraint} \quad	g(x) = b +a_{1}x_{1}^{2} + a_{2}x_{1}\ldots + a_{2n-1}x_n^{2} +a_{2n}x_n
		\label{quadraticeq}
		\end{equation}

or combination of them. We also consider various numbers of these constraints in this study. Here, $x_{1},x_{2} \dots ,x_{n}$ are the variables from Equation \ref{eq:f} and  $a_{1},a_{2} \dots ,a_{n}$ are coefficients within the lower and upper bounds ($l_{c}, u_{c}$). In our research, we construct constrained problems where the optimum of the experimented unconstrained problem is feasible. We use quadratic function as the form of Equation \ref{quadraticeq} (univariate) since it is more popular in recent constrained problem benchmarks. Also, the influence of each $x_{n}$s can be analysed
 independently (exponent 2).
The optimum of these problems is $x^*=(0, \ldots, 0)$ and we ensure that this point is feasible by requiring $b \leq0$, when evolving the constraints.

\subsection{Constraints Features}
In this paper, we study a set of statistic based features that leads to generating hard and easy problem instances. These features are discussed as follows:\\

\begin{itemize}

	\item \textbf{Constraint Coefficients Relationship:} It is likely that the statistics such as standard deviation, population standard deviation and variance of the constraints coefficients can represent the constraints influences to problem difficulty. These constraint coefficients are $(b, a_{1}, a_{2},\dots,a_{n})$ in Equation \ref{lineareq} and \ref{quadraticeq}.
	
	\item \textbf{Shortest Distance: }This feature is related to the shortest distance between the objective function optimum and constraint. In this paper, the shortest distance to the known optimum from each constraint and their relations to each other is discussed. 
	To find the shortest distance of optimum point $(x_{1},x_{2},\dots,x_{n})$ to the linear constraint hyperplane ($a_{1}x_{1}+a_{2}x_{2}+ \dots a_{n}x_{n}+ b=0$) we use Equation \ref{eq:distance}. also, for quadratic constraint hyperplane ($a_{1}x_{1}^{2} +a_{2}x_{1}\ldots +a_{(2n-1)}x_n^{2} +a_{2n}x_n +b =0$) we need to find the minimum of Equation \ref{eq:distance_quad}.

	\begin{equation}
	\label{eq:distance}
	d_{\bot} = \frac{a_{1}x_{01}+a_{2}x_{02}+ \dots a_{n}x_{0n}+ b}{\sqrt{{a_{1}}^2 +{a_{2}}^2+ \dots +{a_{n}}^2 }} 
	\end{equation}
	
	\begin{equation}
	\label{eq:distance_quad}
	d_{\bot} = \sqrt{(x_{1}-x_{01})^2 +(x_{2}-x_{02})^2+\dots +(x_{n}-x_{0n})^2 }
	\end{equation}
	where $d_{\bot}$ in Equation \ref{eq:distance_quad} is the distance from a point to a quadratic hyperplane. Minimizing the distance squared ($d_{\bot}^2$) is equivalent to minimizing the distance $d_{\bot}$.
	
\item \textbf{Angle: } This feature describes the angle of the constraints hyperplanes to each other. It is assumed that the angle between the constraints can influence problem difficulty. To calculate the angle between two linear hyperplanes, we need to find their normal vectors and angle between them using the following equation:
\begin{equation}
\label{eq:angle}
\theta = \arccos \frac {n_{1} \cdot n_{1}}{|n_{1}||n_{2}|}
\end{equation}
where $n_{1}$,$n_{2}$ are normal vectors for two hyperplanes. Also, the angle between two quadratic constraints is the angle between two tangent hyperplanes of their intersection. Then, the angle between these tangent hyperplanes can be calculated by Equation \ref{eq:angle}.

\item \textbf{Number of Constraints: } Number of constraints plays an important role in problem difficulty. In this research, number of constraints and their effects to make easy and hard problem instances is analysed.

\item \textbf{Optimum-local Feasibility Ratio: } Although the global feasibility ratio is important to find the initial feasible point, it should not affect the convergence rate during solving the problem. So, in this research, he feasibility ratio of generated COP is calculated by choosing $10e6$ random points within the vicinity of the optimum in search space and the ratio of feasible points to all chosen ones is reported. In our experiment, the vicinity of optimum is equivalent to 1/10 of boundaries from optimum for each dimension.	
	
\end{itemize}

	\section{Experimental Analysis}
	We now analyse the features of constraints (linear, quadratic and their combination) for easy and hard instances. We generate these instances for ($\epsilon$DEg) algorithm using well known objective functions.
	In our experiments, we generate two sets of hard and easy problem instances. Due to stochastic nature of evolutionary algorithms, for each number of constraints we perform 30 independent runs for evolving easy and hard instances. We set the evolving algorithm (DE) generation number to 5000 for obtaining the proper easy and hard instances. The other parameters of evolving algorithm are set to population size = $40$, crossover rate = $0.5$, scaling factor = $0.9$ and $FEN_{max}$ is $300,000$. Values for these parameters have been obtained by optimising the performance of the evolving algorithm in order to achieve the more easier and harder problem instances. For  ($\epsilon$DEg) algorithm, its best parameters are chosen based on \cite{takahama2010constrained}. The ($\epsilon$DEg) algorithm parameters are considered as: generation number = $1500$, population size = $40$, crossover rate = $0.5$, scaling factor = $0.9$. Also, the parameters for e-constraint method are set to control generation ($Tc$) = $1000$, initial $e$ level ($q$) = $0.9$, archive size = $100n$ ($n$ is dimension number), gradient-based mutation rate ($Pg$) = 0.2 and number of repeating the mutation ($Rg$) = 3.
	
	\subsection{Analysis for Linear Constraints }
	In order to focus only on constraints, we carry out our experiments on various well-known objective functions. These functions are considered as: Sphere (bowl shaped), Ackley (many local optima), Rosenbrock (valley shaped) and Schaffer (many local minima) (see \cite{hansen2010real}). The linear constraint is as the form of Equation \ref{lineareq} with dimension ($n$) as 30 and all coefficients $(a_{n})$s and $b$s are within the range of $[-5,5]$. Also, number of constraints is considered as $1$ to $5$. To discuss and study some features such as shortest distance to optimum, we assume that zero is optimum (all $b$s should be negative). We used  ($\epsilon$DEg) algorithm as solver to generate more easy and hard instances. 
	
\begin{figure}

	\centering		
	\includegraphics[scale=0.6]{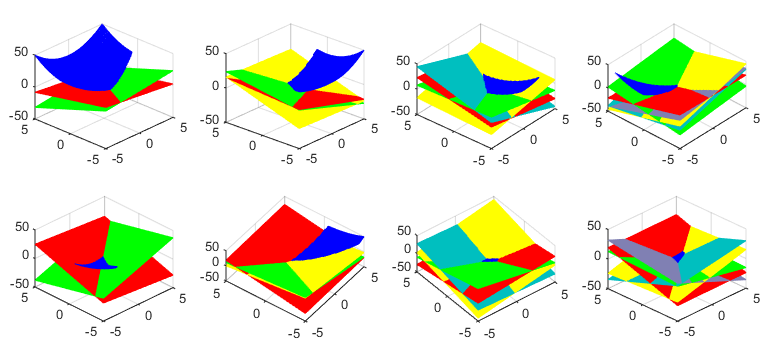}
	\caption{
	Easy (first row) and hard (second row) instances for 1 to 5 number of linear constraints using $\epsilon$DEg (2 dimension). The dark blue hyperplane is the feasible solution}
	\label{fig:plotlinear}
	\end{figure}

	To illustrate our investigation, we plot a 2 dimension Sphere function with 2 to 5 linear constraints in Figure \ref{fig:plotlinear}. It is obvious that the first row (easy) instances have higher feasibility ratio than the second row (hard).
	
	In the following we will present our findings based on various features for linear constraints (for each dimension).\\

	\begin{figure}

		\centering		
		\includegraphics[scale=0.55]{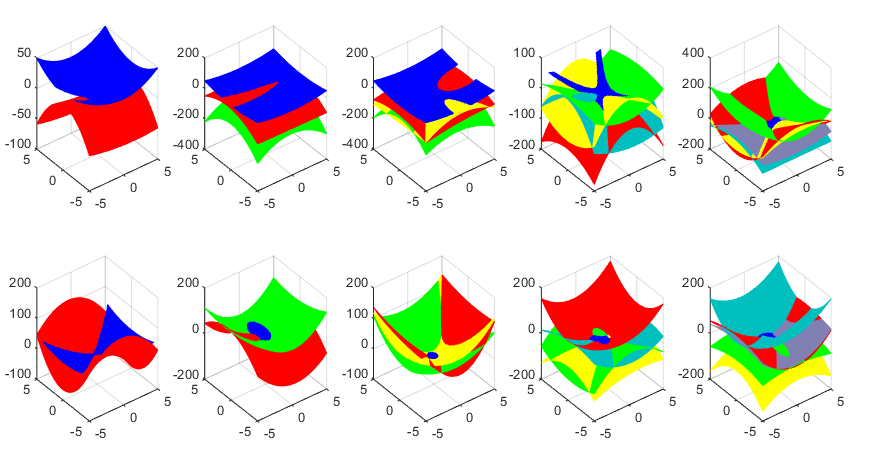}
		\caption{
			Easy (first row) and hard (second row) instances for 1 to 5 number of quadratic constraints using $\epsilon$DEg (2 dimension). The dark blue hyperplane is the feasible solution}
	\end{figure}

	Figure \ref{fig:box_plot_std} shows some evidence about linear constraints coefficients relationship such as standard deviation. It is obvious that there is a systematic relationship between the standard deviation of linear constraint coefficients and problem difficulty. The box plot (see Figure \ref{fig:box_plot_std}) represents the results for easy and hard instances using Sphere, Ackley, Rosenbrock and Schaffer objective function for  ($\epsilon$DEg) algorithm (solver). As it is observed, the standard deviation for coefficients in each constraint (1 to 5) for easy instances are lower than hard ones. Both these coefficient values can be a significant role to make a constrained problem harder or easier to solve. Also, interestingly, all different objective functions follow the same pattern.\\

	Figure \ref{fig:box_plot_dis} represents variation of shortest distance to optimum feature for easy and hard instances using  ($\epsilon$DEg) algorithm. The lower value means the higher distance from optimum. This means, the linear hyperplanes in easy instances are further from optimum. Based on results, there is a strong relationship between problem hardness and shortest distance of constraint hyperplanes to optimum.  In other word, this feature is contributing to problem difficulty. As expected, all objective functions follow the same systematic relationship between their feature and problem difficulty. This means, this feature can be used as a proper source of knowledge for predicting problem difficulty.\\

	The angle between linear constraint hyperplanes feature shows relationship between the angle and problem difficulty (see Table \ref{table:anglelinear}). As it is observed in this table, the angle between constraints in easier instances are less than higher ones. So, this feature is contributing in problem difficulty.

\begin{table}
	
	\caption {The angle feature for Sphere objective function}
	{
	\hspace*{-1.2cm}
		\centering
		\resizebox{1.2\columnwidth}{!}{%
			
			\begin{tabular}{|c|c|c|c|c|c|c|c|c|c|c|c|}
				
				\hline \hline
				& Cons 1,2 & Cons 1,3 & Cons 1,4 & Cons 1,5 & Cons 2,3 & Cons 2,4 & Cons 2,5 & Cons 3,4 & Cons 3,5 & Cons 4,5 \\ \hline
				DE Easy & 15  & 17  &   25  & 21  & 32 & 27 & 41 &   47  & 45  & 43\\ \hline
				DE Hard & 45  & 51 &   63  & 59  & 62 & 73  & 76  &   69  & 79  & 86\\ \hline
				
			\end{tabular}}
			\setlength{\tabcolsep}{5em}
			
			\label{table:anglelinear}
		}
	\end{table}

	Table \ref{table:Fenlin} explains the variation of number of constraints feature group. It is shown that the problem difficulty (required FEN for easy and hard instances) has a strong systematic relationship with number of constraints for the experimented algorithm. 
	
	To calculate the optimum-local feasibility ratio, $10e^{6}$ points are generated within the vicinity of optimum (zero in our problems). Later, the ratio of feasible points to all generated points are investigated for easy and hard instances. Results point out that increasing number of linear constraints, decreases the feasibility ratio for experimented algorithms (see Table \ref{table:feasibilityratiolin}).\\
	
In summary the variation of feature values over the  problem difficulty is more prominent in some of them than the other groups of features. Features such as, coefficients standard deviation, shortest distance, angle between constraints, number of constraints and feasibility ratio exhibit a relationship to problem hardness. This relationship is stronger for some features.

\begin{table}
	\hspace*{-0.4cm}
	\resizebox{0.42\columnwidth}{!}{%
		\parbox{.49\linewidth}{
			\centering
			
			\caption{The FEN for linear constraints}
			
			\begin{tabular}{|P{2.8cm}|P{1.5cm}|P{1.5cm}|}
				
				\hline \hline
				Constraint - Function 
				& Easy Instance & Hard Instance \\ \hline
				1 c Sphere& 25.6K  &  91.2K   \\ \hline
				2 c Sphere& 28.9K &  93.4K  \\ \hline
				3 c Sphere& 32.4K &  98.3K    \\ \hline
				4 c Sphere& 34.2K & 104.2K   \\ \hline		
				5 c Sphere& 35.5K & 123.2K     \\ \hline
				1 c Ackley& 65.2K &  232.1K    \\ \hline
				2 c Ackley& 69.3K &  243.7K  \\ \hline
				3 c Ackley& 74.2K &  265.4K   \\ \hline
				4 c Ackley& 86.4K & 271.3K   \\ \hline		
				5 c Ackley& 92.3K & 277.2K    \\ \hline
				1 c Rosenbrock& 32.8K  &  145.2K    \\ \hline
				2 c Rosenbrock& 35.9K &  153.3K    \\ \hline
				3 c Rosenbrock& 34.5K &  167.9K     \\ \hline
				4 c Rosenbrock& 42.2K & 172.4K   \\ \hline		
				5 c Rosenbrock& 48.3K & 176.8K  \\ \hline
				1 c Schaffer& 84.8K  &  247.1K   \\ \hline
				2 c Schaffer& 87.9K &  259.1K  \\ \hline
				3 c Schaffer& 93.5K &  280.3K  \\ \hline
				4 c Schaffer& 103.2K & 293.8K  \\ \hline		
				5 c Schaffer& 112.4K & 297.4K   \\ \hline
				
			\end{tabular}
			\label{table:Fenlin}
		}
		
	}
	\hfill
	\resizebox{0.51\columnwidth}{!}{%
		\parbox{.6\linewidth}{
			\centering
			\caption{The FEN for quadratic constraints}
			\begin{tabular}{|P{2.8cm}|P{1.5cm}|P{1.5cm}|}
				
				\hline \hline
				Constraint - Function 
				& Easy Instance & Hard Instance \\ \hline
				1 c Sphere& 24.2K  &  129.3K \\ \hline
				2 c Sphere& 25.3K  &  132.6K  \\ \hline
				3 c Sphere&  27.9K &  136.2K  \\ \hline
				4 c Sphere&  34.1K &  141.2K    \\ \hline	
				5 c Sphere&  38.7K &  149.3K    \\ \hline
				1 c Ackley&  68.4K &  228.3    \\ \hline
				2 c Ackley&  72.9K &  232.5K   \\ \hline
				3 c Ackley&  84.5K &  239.6K   \\ \hline
				4 c Ackley&  95.3K &  247.9K   \\ \hline	
				5 c Ackley&  98.1K &  251.9K    \\ \hline
				1 c Rosenbrock&  31.4K &  173.2K \\ \hline
				2 c Rosenbrock&  32.45K &   182.3K  \\ \hline
				3 c Rosenbrock&  42.5K &   190.6K \\ \hline
				4 c Rosenbrock&   52.7K &  192.8K  \\ \hline		
				5 c Rosenbrock&  71.1K &  213.4K \\ \hline
				1 c Schaffer& 91.3K  &  278.9K   \\ \hline
				2 c Schaffer&  94.9K &  283.1K \\ \hline
				3 c Schaffer& 103.7K  &  289.3K   \\ \hline
				4 c Schaffer&  114.1K &  296.1K   \\ \hline		
				5 c Schaffer&  123.4 &  300k   \\ \hline
				
			\end{tabular}
			\label{table:Fenquad}
		}
	}

\end{table}

\subsection{Analysis for Quadratic Constraints}
In this section, we carry out our experiments on quadratic constraints. We use various objective functions, dimension and coefficient range similar to linear analysis. In the following the group of features are studied for easy and hard instances using quadratic constraints.\\

\begin{figure*}[t]
	\hspace*{-2.5cm}
	\centering
	\includegraphics[width=1.3\textwidth]{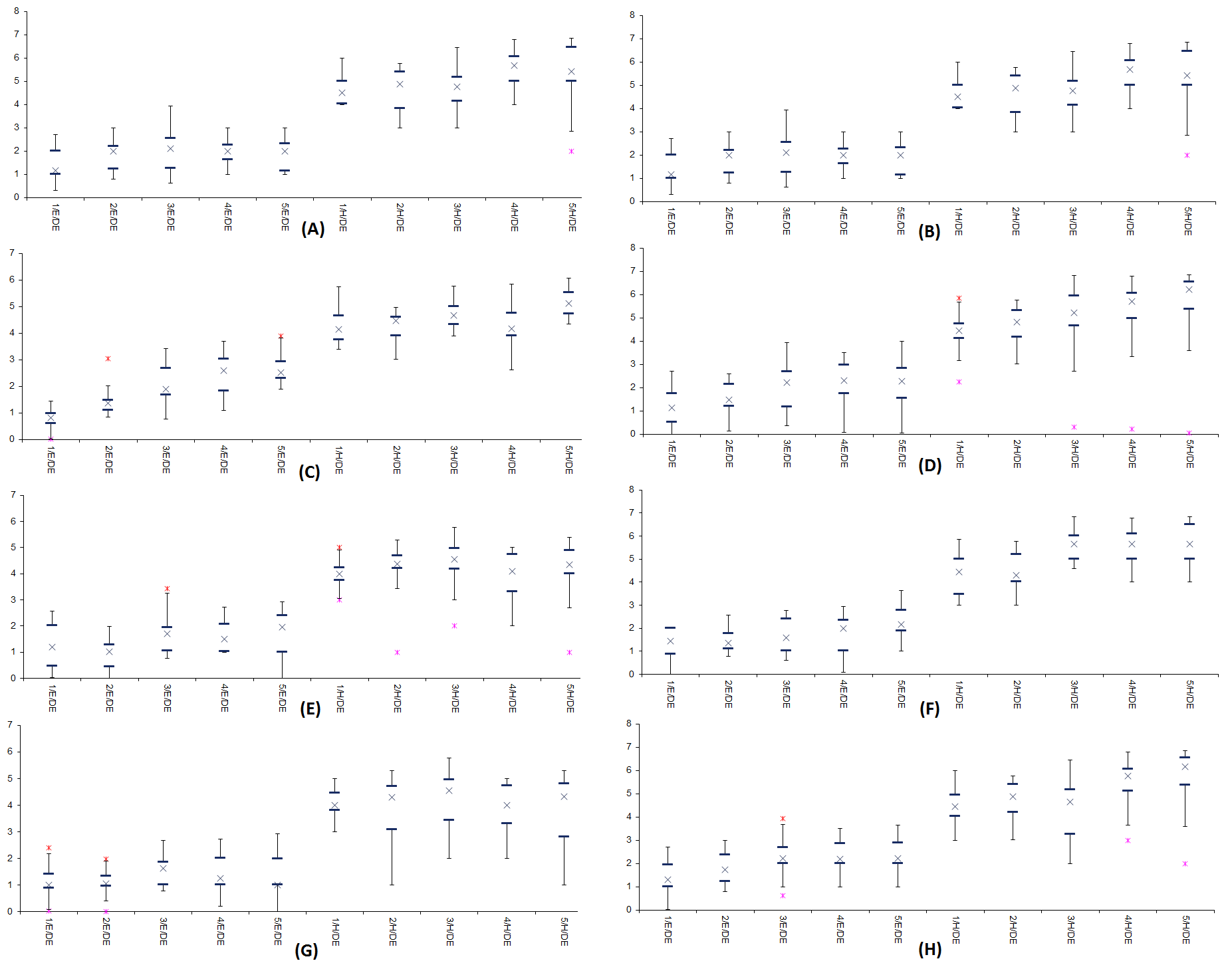}
	\caption{Box plot for standard deviation of coefficients in linear (A,C,E,G) and quadratic (B,D,F,H) constraints for Sphere (A,B), Ackley (C,D), Rosenbrok (E,F) and Schaffer (G,H). Each sub figure includes 2 sets of hard (H) and Easy (E) instances with 1 to 5 constraints using algorithms (a/b/c denotes a: constraint number, b: easy/hard instances and c:algorithm).} 
	\label{fig:box_plot_std}
\end{figure*}

Observing the Figure \ref{fig:box_plot_std}, we can identify the relationship of quadratic coefficients and their ability to make problem hard or easy. Based on the experiments, quadratic coefficients  has the ability to make problems hard or easier for algorithms. In other words, in each constraint, the quadratic coefficients (within the quadratic constraint) are more contributing to problem difficulty than linear coefficients (see Equation \ref{quadraticeq}). Figure \ref{fig:box_plot_std} shows the standard deviation of quadratic coefficients for easy and hard COPs. As shown, the standard deviation of quadratic coefficient in 1 to 5 constraints in easy instances are less than harder one. In contrast to quadratic coefficients, our experiments show there is no systematic relationship between the linear coefficient in quadratic constraints and problem hardness. In other words, quadratic coefficients are more contributing than linear ones in the same quadratic constraint.\\

\begin{figure*}[t]
	\centering
		\hspace*{-2.5cm}
	\includegraphics[width=1.3\textwidth]{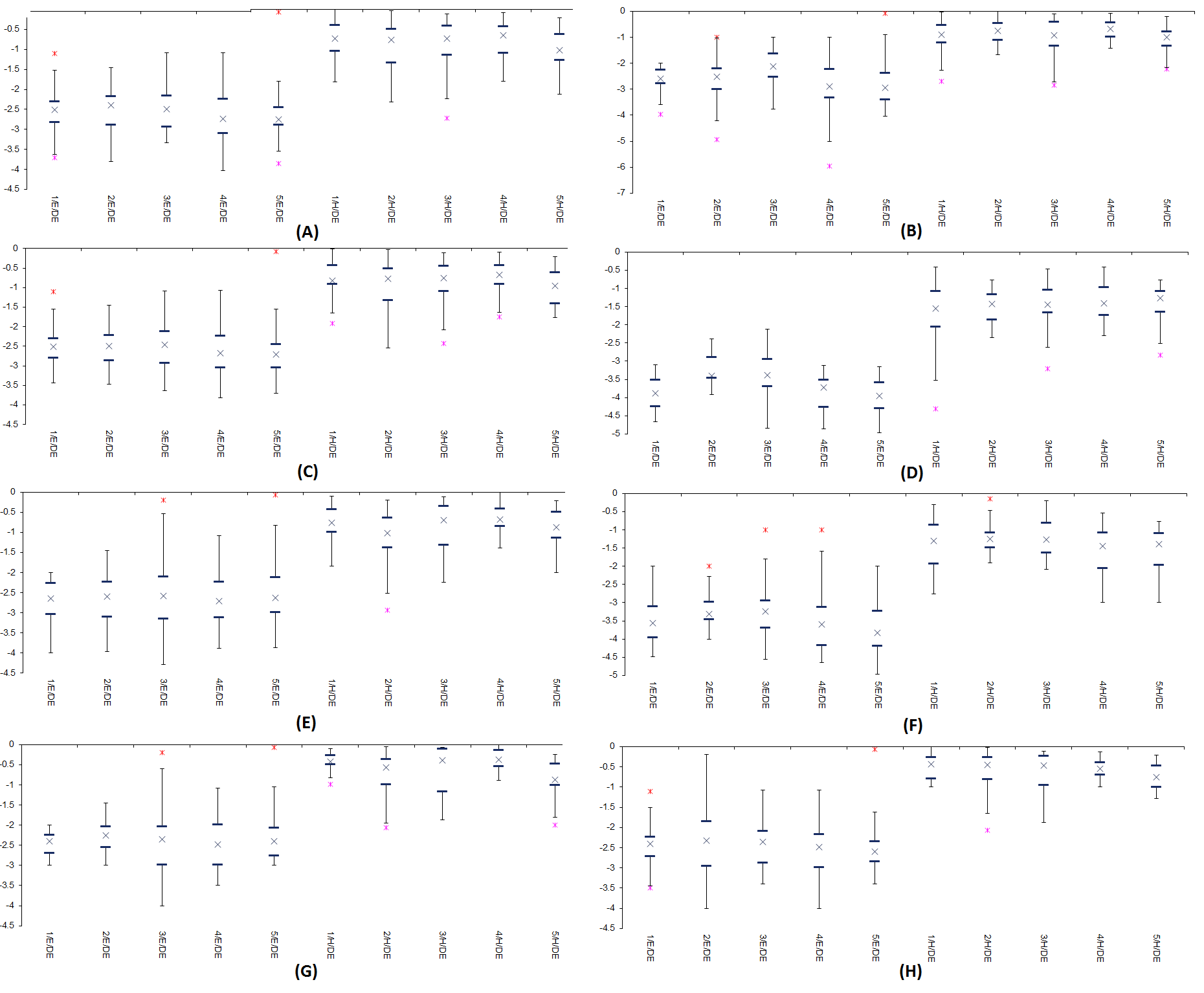}
	\caption{Box plot for the shortest distance to optimum of linear (A,C,E,G) and quadratic (B,D,F,H) constraints for Sphere (A,B), Ackley (C,D), Rosenbrok (E,F) and Schaffer (G,H). Each sub figure includes 2 sets of hard (H) and Easy (E) instances with 1 to 5 constraints using DE algorithm (a/b/c denotes a: constraint number, b: easy/hard instances and c:algorithm).} 
	\label{fig:box_plot_dis}

\end{figure*}

Box plots shown in Figure \ref{fig:box_plot_dis} represent the shortest distance of a quadratic constraint hyperplanes to optimum. As it is observed, harder instances have constraint hyperplanes closer to optimum than easier ones. The lower values in these box plots means closer to optimum.	Calculating the angles between constraints do not follow any systematic pattern and there is no relationship between angle feature and problem difficulty for quadratic constraints. 
We also study the number of quadratic constraints feature. As it is shown in Table \ref{table:Fenquad}, number of quadratic constraints is contributing to problem difficulty. It is obvious that increasing number of quadratic constraints makes a problem harder to solve (increases FEN).
As observed in Table \ref{table:feasibilityratioquad}, investigations on feasibility ratio show that increasing number of constraint decreases the problem optimum-local feasibility ratio for easy and hard instances respectively.
As it is observed, some group of features are more contributing to problem difficulty than the others. It is shown that angle feature does not follow any systematic relationship with problem hardness for experimented algorithm for quadratic constraints. On the other hand standard deviation, feasibility ratio and number of constraints are more contributing for $\epsilon$DEag.

\begin{table}
	\centering
		
	\caption{Optimum-local feasibility ratio of search space near the optimum for 1,2,3,4 and 5 linear constraint}

	\begin{tabular}{|c|c|c|}
						
			\hline \hline
			& DE Easy & DE Hard \\ \hline
			1 cons & 42\% & 7\% \\ \hline
			2 cons & 32\% &  6\%  \\ \hline
			3 cons & 22\% & 4\%   \\ \hline
			4 cons & 17\% & 3\%    \\ \hline		
			5 cons & 11\% & 2\%    \\ \hline
			\end{tabular}
				
				\label{table:feasibilityratiolin}
\end{table}
\begin{table}
	\centering
		\caption{Optimum-local feasibility ratio of search space near the optimum for 1,2,3,4 and 5 quadratic constraint}

	\begin{tabular}{|c|c|c|}
						
				\hline \hline
				& DE Easy & DE Hard \\ \hline
				1 cons & 36\% & 11\% \\ \hline
				2 cons & 27\% & 7\% \\ \hline
				3 cons & 12\% & 4\%   \\ \hline
				4 cons & 11\% & 3\%   \\ \hline		
				5 cons & 8\% & 2\%    \\ \hline
			\end{tabular}
				
				\label{table:feasibilityratioquad}
			\end{table}
		
\begin{table}
		\centering
				\caption{The FEN for combined constraints using Sphere objective function}

					\begin{tabular}{|c|c|c|}
						\hline \hline
						& DE Easy & DE Hard \\ \hline
						1 Lin 4 Quad & 22.4K  &  97.5K  \\ \hline
						2 Lin 3 Quad & 17.5K  &  95.1K  \\ \hline
						3 Lin 2 Quad & 16.5K  &  94.2K  \\ \hline						
						4 Lin 1 Quad & 14.1K  &  91.4K    \\ \hline
					\end{tabular}
				
				\label{table:feasibilityratioquadlin}

	\end{table}

	\subsection{Analysis for Combined Constraints}
	In this section, we consider the combination of linear and quadratic constraints. The generated COPs have different numbers of linear and quadratic constraints (5 constraints). The obtained results show the higher effectiveness of quadratic constraints. In other words, these constraints are more contributing to problem difficulty than linear ones. By analysing the various number of constraints (See Table \ref{table:feasibilityratioquadlin}) we can conclude that required FEN for sets of constraints with more quadratic ones is higher than sets with more linear constraints. This relationship holds the pattern for both easy and hard instances.\\
	
	In summary it is observed that the variation of linear and quadratic constraint coefficients over the problem difficulty is more contributing for some group of features. Considering quadratic constraints only, it is obvious that some features such as angle do not provide useful knowledge for problem difficulty. In general, this experiments point out the relationship of the various constraint features of easy and hard instances with the problem difficulty while moving from easy to hard ones. This improves the understanding of the constraint structures and their ability to make a problem hard or easy for a specific group of evolutionary algorithms.

	\section*{Conclusions}
	In this paper, we performed a feature-based analysis on the impact of sets of constraints (linear, quadratic and their combination) on performance of well-known evolutionary algorithm ($\epsilon$DEag). Various features of constraints for easy and hard instances have been analysed to understand which features contribute more to problem difficulty. The sets of constraints have been evolved using an evolutionary algorithm to generate hard and easy problem instances for $\epsilon$DEag. Furthermore, the relationship of the features with the problem difficulty have been examined while moving from easy to hard instances. Later on, these results can be used to design an algorithm prediction model.

\section*{Acknowledgements}
Frank Neumann has been supported by ARC grants DP130104395 and DP140103400.

\end{document}